\documentclass[11pt]{article}

\usepackage[final]{acl}

\usepackage{times}
\usepackage{latexsym}

\usepackage[T1]{fontenc}

\usepackage[utf8]{inputenc}

\usepackage{microtype}

\usepackage{inconsolata}

\usepackage{graphicx}

\usepackage{booktabs}
\usepackage{float}
\usepackage{multirow}

\title{Using an LLM to Tackle Difficult Cases in NER for Nepali Tweets}

\author{Claire Wang \\
  Brandeis University}

\begin{document}
\maketitle
\begin{abstract}
Named entity recognition (NER) is challenging for social media data in a low-resource language. LLMs often lag behind pretrained language models (PLMs) in NER performance, but they show promise in handling cases that PLMs struggle with. Using a dataset of Nepali tweets, we split the test set into ``easy'' and ``hard'' cases using token probabilities from a fine-tuned pretrained language model (PLM). We then evaluate an LLM on the hard test cases using 16 different prompt configurations. We find that selecting examples based on semantic similarity and increasing the number of examples improves F1 score by almost 10 points from the baseline configuration. Although the LLM underperforms the PLM on the full and easy test sets, it surpasses the PLM on the hard test set, resulting in an improved score when the two models are combined.
\end{abstract}

\section{Introduction}

\begin{figure}[t]
  \includegraphics[width=\linewidth]{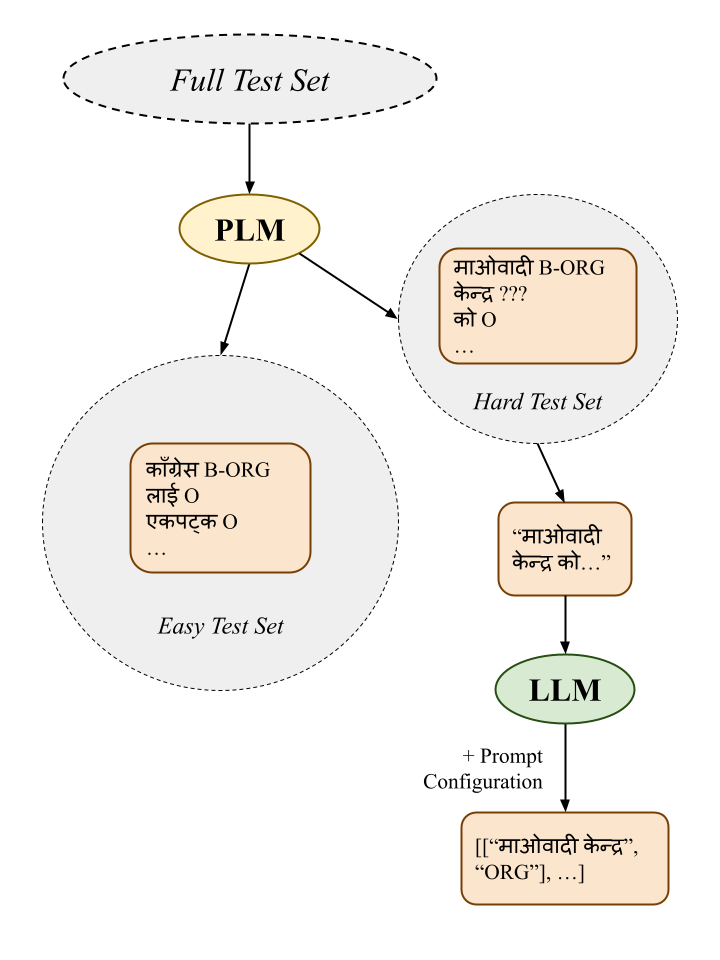}
  \caption{Illustration of our framework. The pretrained language model sorts test tweets as easy or hard based on token probabilities; hard tweets are sent to the LLM for evaluation using each prompt configuration.}
  \label{fig:framework}
\end{figure}

Named entity recognition (NER) is especially challenging for certain types of data. This includes user-generated text such as social media posts, which are often noisy, lacking in context, and full of unseen entities \cite{ugt-lit-review}. NER in low-resource languages can also be difficult; there is less labeled data that can be used for training, and some languages have morphological or structural characteristics that systems developed for other languages do not account for \cite{lr-lit-review}. Even models specialized for a particular language may struggle with code-switching between languages, which is common in social media data \cite{aguilar-etal-2018-named}.

Encoder-only transformer-based models, which we will refer to as pretrained language models (PLMs), have achieved state-of-the-art results for NER in many domains \cite{keraghel2024recentadvancesnamedentity}. So far, large language models (LLMs) have not had the same level of success: despite their impressive capabilities, they are designed for text generation and NER is a sequence labeling task \cite{wang-etal-2025-gpt}.

Even if LLMs generally cannot outperform PLMs in NER, some papers have found that LLMs do better than PLMs on challenging cases. This serves as justification for using the two models together, with the LLM complementing the PLM by handling cases that the PLM is less certain about \cite{ma-etal-2023-large, LinkNER}. Building on this idea, we use Nepali Twitter data to investigate the following research questions:

\textbf{RQ1:} \textit{Is token probability an effective way of identifying cases that are hard for a PLM but easier for an LLM?}

To effectively use a PLM and an LLM in conjunction, we need to determine when to trust the predictions of each model. If the LLM performs worse overall, this is a key decision, and since the LLM has higher inference cost, it is beneficial to not use it for more predictions than necessary. Token probability is an easy-to-calculate estimate of a PLM's confidence in its predictions, so it serves as a good starting point.

\textbf{RQ2:} \textit{How do techniques that have shown promise in improving NER performance for LLMs perform on noisy data in a low-resource language?}

Techniques such as including annotation guidelines or entity type definitions in the prompt, few-shot learning with examples similar to the input, and asking the model to verify its own predictions have been shown to improve LLM performance on some NER datasets \cite{sainz2024gollie, huang-2025-guidener, wang-etal-2025-gpt}. However, more research is necessary to determine whether this holds across languages and for noisy data such as tweets.

\textbf{RQ3:} \textit{Can a multilingual LLM outperform a monolingual PLM on difficult cases, especially ones containing code-switching?}
 
Although LLMs are often trained on many languages, it is unclear if they can perform equally well in all of them, especially for specialized tasks such as NER. We expect the PLM to have a better ``understanding'' of Nepali, but \citet{Niraula_Chapagain_2023} observed that code-switching was a prominent source of error when evaluating PLMs on the same dataset we use. We hypothesize that our LLM will outperform our PLM on code-switched examples. Previous work suggests that the LLM will also perform better on hard cases overall, but these findings were based on English data, so it remains to be seen whether the same pattern holds for a low-resource language such as Nepali.

\section{Related Work}
Many publications have investigated the use of LLMs for NER \cite{keraghel2024recentadvancesnamedentity}. One notable example is GPT-NER \cite{wang-etal-2025-gpt}, which implemented several techniques that we also use (self-verification and example selection based on semantic similarity).

Most LLM NER research has focused on high-resource languages such as English, but \citet{Neupane_2025} and \citet{yadav-etal-2025-prompt} use Nepali news data. Both had F1 scores more than 10 points below those previously achieved by PLMs on their respective datasets.

\citet{ma-etal-2023-large} found that LLMs underperformed PLMs in general and on easy cases but outperformed them on hard cases. They thus developed a framework where the PLM made initial predictions but passed entities with a low confidence score to the LLM to be re-evaluated. They used FewNERD, an English Wikipedia dataset, and achieved a 1.1\% increase in F1 (comparing their best PLM + LLM score to their best PLM score).

\citet{LinkNER} employed a similar strategy of using a PLM for the first round of predictions and turning to an LLM for entities where the PLM was less certain. They used English data from social media as well as less noisy domains, achieving a 3.41\% F1 increase on the English Twitter dataset TweetNER.

To our knowledge, no previously published studies have explored the PLM + LLM strategy for data in a low-resource language. We additionally contribute to the current literature by thoroughly testing combinations of several prompting techniques (\citet{ma-etal-2023-large} and \citet{LinkNER} experiment with some prompting/shot number variations, but not as extensively). In addition, both of these papers present the NER task to the LLM as a simple multiple-choice selection, where it only needs to classify the type of a span that has already been identified by the PLM as a potential entity mention. We diverge from this by only providing the LLM with the input tweet, forcing it to identify mention boundaries on its own. This allows the model to potentially correct boundary errors made by the PLM, and it gives us a more realistic understanding of the LLM's NER capabilities. 

\section{Methodology}

We first ran a PLM on the test set to construct a subset of more difficult test data. We then evaluated an LLM on this subset using different prompt configurations.

\begin{table*}
\centering
\begin{tabular}{lrrrrrrr}
\toprule
& & \multicolumn{6}{c}{\textbf{\# of Mentions}} \\ \cmidrule(lr){3-8}
\textbf{Data Split} & \textbf{\# of Tweets} & \textbf{Total} & \textbf{PER} & \textbf{LOC} & \textbf{ORG} & \textbf{DATE} & \textbf{EVENT} \\
\midrule
Train & 5366 & 3463 & 1061 & 923 & 782 & 663 & 34 \\
Full Test & 2299 & 1503 & 444 & 389 & 356 & 286 & 28 \\
Easy Test & 2085 & 1113 & 328 & 304 & 261 & 208 & 12 \\
Hard Test & 214 & 390 & 116 & 85 & 95 & 78 & 16 \\
\bottomrule
\end{tabular}
\caption{Dataset splits with the number of tweets and mentions (both total and by entity type) in each split.}
\label{tab:data-splits}
\end{table*}

\subsection{Data Preparation}
We used the DanfeNER dataset \cite{Niraula_Chapagain_2023}, a human-annotated collection of Nepali tweets containing five entity types: Person (PER), Location (LOC), Organization (ORG), Date (DATE), and Event (EVENT). The original data files contained incorrectly formatted blocks of non-annotated tweets and inconsistencies between annotations in the two versions of the test file (one in BIO format and one with mentions listed as character spans). We removed the non-annotated tweets and corrected the inconsistencies, using the BIO file as the source of truth.

\subsubsection{Construction of Hard Test Set}
To identify tweets that were challenging for the PLM, we trained it on the train set and evaluated it on the test set. We approximated the probabilities for each predicted token using softmax and selected all tweets containing a token with a prediction probability less than 90\%.

Table~\ref{tab:data-splits} displays statistics for all dataset splits, where the hard test set was created through the process described above and the easy test set is composed of all test tweets not in the hard test set. 

\subsubsection{Data Format}
The PLM was trained on and returned tokens labeled using the BIO tagging scheme. Transition errors (O $\to$ I and B $\to$ I for different entity types) were repaired using the \texttt{conlleval} method of the \texttt{seqscore} tool \cite{palen-michel-etal-2021-seqscore} before calculating F1 scores.

Since LLMs struggle with BIO format \cite{wang-etal-2025-gpt}, we provided the LLM with the tweet as a single string (non-tokenized) and instructed it to return predicted mentions  in the format \texttt{[["entity1", "type1"], ["entity2", "type2"], ...]}, using the Pydantic library to enforce that the model returned a list of list of strings. Examples in the prompt were formatted the same way.

Formatting errors were rare. There was one occurrence of the entity and type being in the same string (\texttt{["entity, type"]}). All remaining errors were the model outputting \texttt{[[]]} instead of \texttt{[]}. These were corrected through a find-and-replace.

\subsection{Prompt Engineering}
We examined the effect of three variables on the prompt: inclusion of entity type definitions, number of examples given, and example selection strategy. We then used self-verification on the best-performing configuration. The full text of the prompts is given in Appendix~\ref{app:prompts}.

\subsubsection{Entity Type Definitions}
Entity type definitions were adapted from the annotation guidelines for DanfeNER, which were the same guidelines used for an NER dataset of Nepali news articles, EverestNER \cite{Niraula_Chapagain_2022}. The guidelines were published in English, so our definitions were given in English as well. We either included the definitions in the prompt or listed the entity types without any definitions.

\subsubsection{Examples}

\begin{table*}
\centering
\begin{tabular}{r|rrrr}
\toprule
& \multicolumn{2}{c}{\textbf{Random Examples}} & \multicolumn{2}{c}{\textbf{Similar Examples}} \\
& \textbf{No Definitions} & \textbf{Definitions} & \textbf{No Definitions} & \textbf{Definitions} \\ \midrule
\textbf{0 Examples} & 57.92 & 58.21 & 57.92 & 58.21 \\
\textbf{10 Examples} & 59.22 & 59.27 & 63.77 & 63.22 \\
\textbf{20 Examples} & 61.36 & 61.43 & 66.31 & 65.18 \\
\textbf{40 Examples} & 63.64 & 63.27 & 67.30 & \textbf{67.72} \\ \bottomrule
\end{tabular}
\caption{F1 scores (averaged over 3 runs) obtained by LLM on hard test set under different prompt configurations (number of examples, example selection strategy, and whether entity type definitions were present).}
\label{tab:prompt-configs}
\end{table*}

We used two selection strategies for examples: \textbf{Random}, where we randomly chose examples from the train set, and \textbf{Similarity-Based}, where we chose the examples from the train set that were most semantically similar to the tweet being evaluated. We measured similarity by using Nepali word embeddings\footnote{NPVec1 GloVe, Base + Normalized + Tokenized + Stemmed version: https://github.com/nowalab/nepali-word-embeddings} to obtain a vector representation for each token, averaging the token vectors for each tweet, and computing cosine similarity between each train tweet vector and the test tweet vector.

In both strategies, we selected a distinct set of examples for each test tweet. (For the random examples, this was accomplished by using a unique random seed for each test tweet.) We also ensured that no more than 25\% of the examples contained zero entities.

The example counts we tested were 0, 10, 20, and 40.

\subsection{Self-Verification} 
When performing NER, LLMs tend to overpredict mentions by overconfidently labeling non-entities as entities (\cite{ma-etal-2023-large, wang-etal-2025-gpt}). To address this, after testing all configurations to identify the best-performing one, we had the model verify its own predictions. For each predicted mention, we prompted the model to verify whether it was truly a mention of the predicted type, providing the type definition and the tweet containing the mention. If the model answered "no", we removed the mention to obtain an updated set of predictions with potentially fewer false positives.

\section{Experimental Setup}

For our PLM, we chose the model that achieved the highest scores on DanfeNER in previous experiments \cite{Niraula_Chapagain_2023}. This was \texttt{distilbert-base-nepali},\footnote{https://huggingface.co/Sakonii/distilbert-base-nepali} a version of DistilBERT pre-trained on Nepali text. We used the same training parameters as  \citet{Niraula_Chapagain_2023}: 10 epochs, a training batch size of 10, and a learning rate of 0.0001.

For our LLM, we used Gemma 4 26B-A4B with temperature 0. 

All experiments were run on Google Colab with a High-RAM T4 GPU. We ran the full test set experiments (Table~\ref{tab:full-scores}) five times and all other experiments three times. To ensure consistency in our hard test set, we only used the predictions from the first PLM run to construct it.

All reported F1 scores are entity-level and micro-average.

\section{Results}

\begin{figure}[h!]
    \centering
    \includegraphics[width=\linewidth]{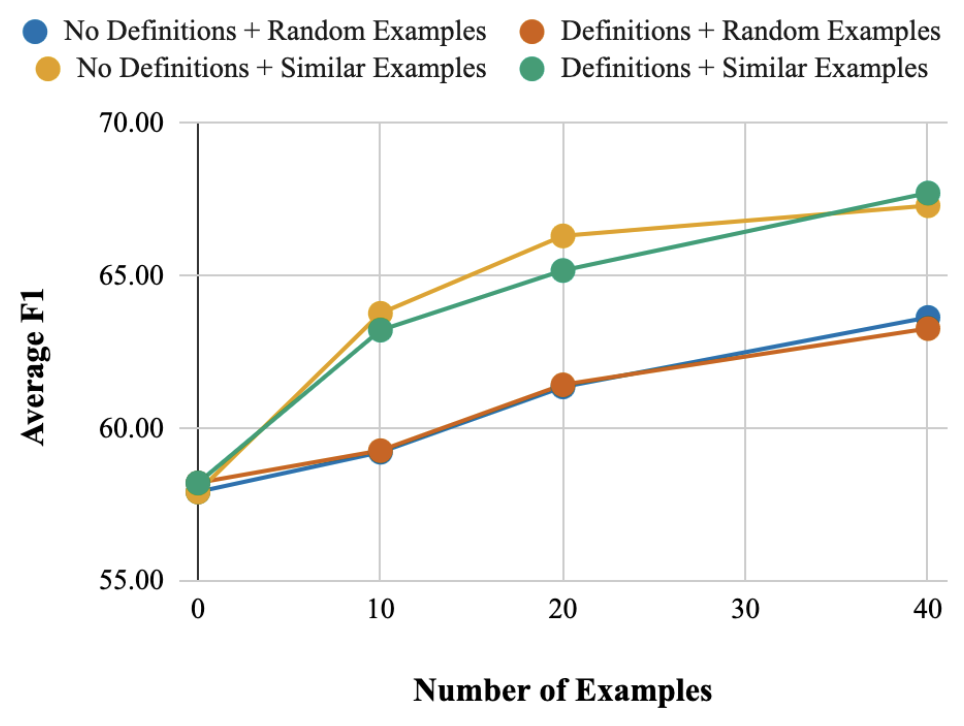}
    \caption{Graph of data in Table~\ref{tab:prompt-configs}.}
    \label{fig:placeholder}
\end{figure}

\begin{table}[h!]
\centering
\begin{tabular}{lrrr}
\toprule
\textbf{Model} & \textbf{P} & \textbf{R} & \textbf{F1} \\
\midrule
LLM Initial & 64.14 & \textbf{71.71} & \textbf{67.72} \\
LLM After SV & \textbf{65.38} & 70.17 & 67.69 \\ \bottomrule
\end{tabular}
\caption{Precision, recall, and F1 scores (averaged over 3 runs) for the best-performing LLM configuration before and after self-verification.}
\label{tab:sv}
\end{table}

\begin{table}[h!]
\centering
\setlength{\tabcolsep}{3pt}
\begin{tabular}{lrrr}
\toprule
\textbf{Model} & \textbf{Full Test} & \textbf{Easy Test} & \textbf{Hard Test} \\
\midrule
PLM & 80.78 \small{$\pm$ 0.29} 
    & \textbf{87.02} \small{$\pm$ 0.22}
    & 63.23 \small{$\pm$ 1.06} \\
LLM & 74.06 \small{$\pm$ 0.23} 
    & 77.21 \small{$\pm$ 0.11} 
    & \textbf{64.97} \small{$\pm$ 0.67} \\ \midrule
\small{Combined} & \textbf{81.76} \small{$\pm$ 0.24} & – & – \\ \bottomrule
\end{tabular}
\caption{F1 scores (averaged over 5 runs) with standard error achieved by the PLM and the best-performing LLM configuration (no self-verification) when run on the full test set. The combined score was calculated by combining the PLM's easy test set predictions with the LLM's hard test set predictions.}
\label{tab:full-scores}
\end{table}

Table~\ref{tab:prompt-configs} shows F1 scores on the hard test set achieved by the 16 LLM prompt configurations we tested. The highest-scoring configuration was using 40 examples selected based on similarity along with entity type definitions. We performed a self-verification step on these predictions, achieving the scores shown in Table~\ref{tab:sv}. Finally, we ran this configuration on the entire test set and calculated F1 scores for the full set as well as the easy and hard subsets (Table~\ref{tab:full-scores}).

\begin{table}
\centering
\begin{tabular}{llrrr}
\toprule
& & \multicolumn{3}{c}{\textbf{\% Code-Switched}} \\
\cmidrule(lr){3-5}
\textbf{Data Split} & \textbf{Model} & \textbf{TP} & \textbf{FP} & \textbf{FN} \\
\midrule
\multirow{2}{*}{Easy Test} & PLM & 3.36 & 5.24 & \textbf{7.73} \\ 
& LLM & \textbf{3.82} & \textbf{7.82} & 4.56 \\ \midrule
\multirow{2}{*}{Hard Test} & PLM & 3.86 & \textbf{11.78} & \textbf{11.39} \\
& LLM & \textbf{6.01} & 6.76 & 7.98 \\
\bottomrule
\end{tabular}
\caption{Percentage of true positives, false positives, and false negatives (averaged over 5 runs) that contained code-switching in easy and hard test sets.}
\label{tab:cs}
\end{table}

\section{Discussion}
From the scores in Table~\ref{tab:prompt-configs}, we see that inclusion of entity type definitions seems to have no meaningful impact on performance, even in the zero-shot case. It created less than 1\% difference in all but one case, and scores with definitions were sometimes lower than scores without definitions. This conflicts with previous research that achieved substantial improvement by including entity type definitions \cite{sainz2024gollie, huang-2025-guidener}. 

In contrast, selecting examples based on similarity and including more examples were clearly advantageous. Combining these techniques resulted in an almost 10\% F1 increase from the baseline of zero examples. 

Self-verification increased precision but decreased recall, resulting in almost no change (-0.03\%) to the F1 score (Table~\ref{tab:sv}).

Comparing the scores achieved by the two models on all test data (Table~\ref{tab:full-scores}), the PLM performed better on the full test set and the easy test set, as expected. The LLM achieved a higher score (+1.74\%) on the hard test set, but the small size of the hard test set and the PLM's standard error ($\pm 1.06$) suggest that this effect may not be reliable. This was significantly higher than any of the other standard errors, due to an unusually high F1 score from one run. 

There was an interesting discrepancy between average F1 scores when the LLM was run on only the hard test set (67.72 for the best-performing configuration) and when scores were calculated for the hard test set data after a run on the full test set ($64.97 \pm 0.67$ for the same configuration). Since predictions for each tweet were made through a separate call to the LLM with no shared context, the number of calls made during a run should not have impacted the results.

Since evaluating the LLM on only the hard test set is the more realistic way this system would be used, we used these predictions (and not the Table \ref{tab:full-scores} predictions from the full test set run) when calculating the combined score. We achieved an increase of 0.98\% F1, which is lower than the gains achieved by \citet{ma-etal-2023-large} and \citet{LinkNER} on English data. 

To investigate whether whether code-switching affected performance, we calculated the percentage of true positives, false positives, and false negatives (based on mentions) containing Latin script for each model on the easy and hard test splits (Table \ref{tab:cs}). The LLM had a higher percentage for true positives and a lower percentage for false positives and false negatives, suggesting that it was indeed better at handling difficult code-switched tweets. For the easy test set, the percentages were lower and the PLM/LLM differences were smaller, so it is more difficult to draw conclusions.

\subsection{Future Work}
To achieve a more substantial F1 increase on the full test set using this method, it would be necessary to identify a larger hard set for which the LLM still outperforms the PLM and/or significantly improve LLM performance on the hard set.

To address the first goal, one could easily increase the hard test set size by raising the token probability threshold that we used to divide easy and hard tweets, but it is unclear if the LLM would still outperform the PLM on this set. Using other methods of calculating uncertainty, or accounting for factors such as code-switching, could be promising directions to explore. 

The second goal also has many potential pathways to investigate. Although the inclusion of entity type definitions and the self-verification stage did not improve performance in our experiments, different ways of implementing these concepts could have better results. Continuing to increase the number of examples in the prompt and seeing if scores continue to increase or plateau is another option. Finally, researchers have tried many approaches to using LLMs for NER beyond the ones explored here, and incorporating these could be fruitful as well. 

\section{Conclusion}

Expanding on previous NER research finding that LLMs perform better than PLMs on challenging cases, we divided a test set of Nepali tweets into easy and hard subsets based on token probabilities from a PLM that had been fine-tuned on the train set. In-context learning with a higher number of examples selected based on semantic similarity improved LLM performance on the hard test set, resulting in a small F1 increase on the full test set. Our results showcase the potential of using LLMs for difficult NER cases for noisy data in a low-resource language.

\bibliography{custom}

\begin{thebibliography}{14}
\providecommand{\natexlab}[1]{#1}

\bibitem[{Aguilar et~al.(2018)Aguilar, AlGhamdi, Soto, Diab, Hirschberg, and Solorio}]{aguilar-etal-2018-named}
Gustavo Aguilar, Fahad AlGhamdi, Victor Soto, Mona Diab, Julia Hirschberg, and Thamar Solorio. 2018.
\newblock \href {https://doi.org/10.18653/v1/W18-3219} {Named entity recognition on code-switched data: Overview of the {CALCS} 2018 shared task}.
\newblock In \emph{Proceedings of the Third Workshop on Computational Approaches to Linguistic Code-Switching}, pages 138--147, Melbourne, Australia. Association for Computational Linguistics.

\bibitem[{Esmaail et~al.(2024)Esmaail, Omar, Mohd, Fauzi, and Mansur}]{ugt-lit-review}
Naji Esmaail, Nazlia Omar, Masnizah Mohd, Fariza Fauzi, and Zainab Mansur. 2024.
\newblock \href {https://doi.org/10.1109/ACCESS.2024.3427714} {Named entity recognition in user-generated text: A systematic literature review}.
\newblock \emph{IEEE Access}, 12:136330--136353.

\bibitem[{Huang et~al.(2025)Huang, Xu, Yu, Li, and Lin}]{huang-2025-guidener}
Shizhou Huang, Bo~Xu, Yang Yu, Changqun Li, and Xin~Alex Lin. 2025.
\newblock \href {https://doi.org/10.1609/aaai.v39i23.34591} {Guide{NER}: Annotation guidelines are better than examples for in-context named entity recognition}.
\newblock \emph{Proceedings of the AAAI Conference on Artificial Intelligence}, 39(23):24159–24166.

\bibitem[{Keraghel et~al.(2024)Keraghel, Morbieu, and Nadif}]{keraghel2024recentadvancesnamedentity}
Imed Keraghel, Stanislas Morbieu, and Mohamed Nadif. 2024.
\newblock \href {https://arxiv.org/abs/2401.10825} {Recent advances in named entity recognition: A comprehensive survey and comparative study}.
\newblock \emph{Preprint}, arXiv:2401.10825.

\bibitem[{Ma et~al.(2023)Ma, Cao, Hong, and Sun}]{ma-etal-2023-large}
Yubo Ma, Yixin Cao, Yong Hong, and Aixin Sun. 2023.
\newblock \href {https://doi.org/10.18653/v1/2023.findings-emnlp.710} {Large language model is not a good few-shot information extractor, but a good reranker for hard samples!}
\newblock In \emph{Findings of the Association for Computational Linguistics: EMNLP 2023}, pages 10572--10601, Singapore. Association for Computational Linguistics.

\bibitem[{Neupane et~al.(2025)Neupane, Chapagain, Niraula, and Koirala}]{Neupane_2025}
Sameer Neupane, Jeevan Chapagain, Nobal~B. Niraula, and Diwa Koirala. 2025.
\newblock \href {https://doi.org/10.32473/flairs.38.1.138929} {Generative {AI} for named entity recognition in low-resource language {N}epali}.
\newblock \emph{The International FLAIRS Conference Proceedings}, 38(1).

\bibitem[{Niraula and Chapagain(2022)}]{Niraula_Chapagain_2022}
Nobal Niraula and Jeevan Chapagain. 2022.
\newblock \href {https://doi.org/10.32473/flairs.v35i.130725} {{N}amed {E}ntity {R}ecognition for {N}epali: {D}ata {S}ets and {A}lgorithms}.
\newblock \emph{The International FLAIRS Conference Proceedings}, 35.

\bibitem[{Niraula and Chapagain(2023)}]{Niraula_Chapagain_2023}
Nobal Niraula and Jeevan Chapagain. 2023.
\newblock \href {https://doi.org/10.32473/flairs.36.133384} {Danfe{NER} - {N}amed {E}ntity {R}ecognition in {N}epali {T}weets}.
\newblock \emph{The International FLAIRS Conference Proceedings}, 36(1).

\bibitem[{Palen-Michel et~al.(2021)Palen-Michel, Holley, and Lignos}]{palen-michel-etal-2021-seqscore}
Chester Palen-Michel, Nolan Holley, and Constantine Lignos. 2021.
\newblock \href {https://doi.org/10.18653/v1/2021.eval4nlp-1.5} {{S}eq{S}core: Addressing barriers to reproducible named entity recognition evaluation}.
\newblock In \emph{Proceedings of the 2nd Workshop on Evaluation and Comparison of NLP Systems}, pages 40--50, Punta Cana, Dominican Republic. Association for Computational Linguistics.

\bibitem[{Sainz et~al.(2024)Sainz, Garc{\'\i}a-Ferrero, Agerri, de~Lacalle, Rigau, and Agirre}]{sainz2024gollie}
Oscar Sainz, Iker Garc{\'\i}a-Ferrero, Rodrigo Agerri, Oier~Lopez de~Lacalle, German Rigau, and Eneko Agirre. 2024.
\newblock \href {https://openreview.net/forum?id=Y3wpuxd7u9} {Go{LLIE}: Annotation guidelines improve zero-shot information-extraction}.
\newblock In \emph{The Twelfth International Conference on Learning Representations}.

\bibitem[{Tang et~al.(2024)Tang, Xia, Li, Xu, and Xiong}]{lr-lit-review}
Xiangyan Tang, Dongwan Xia, Yajing Li, Taixing Xu, and Neal~N. Xiong. 2024.
\newblock \href {https://doi.org/10.1007/978-981-99-7161-9_19} {A survey of low-resource named entity recognition}.
\newblock In \emph{The 7th International Conference on Information Science, Communication and Computing}, pages 246--260, Singapore. Springer Nature Singapore.

\bibitem[{Wang et~al.(2025)Wang, Sun, Li, Ouyang, Wu, Zhang, Li, Wang, and Guo}]{wang-etal-2025-gpt}
Shuhe Wang, Xiaofei Sun, Xiaoya Li, Rongbin Ouyang, Fei Wu, Tianwei Zhang, Jiwei Li, Guoyin Wang, and Chen Guo. 2025.
\newblock \href {https://doi.org/10.18653/v1/2025.findings-naacl.239} {{GPT}-{NER}: Named entity recognition via large language models}.
\newblock In \emph{Findings of the Association for Computational Linguistics: NAACL 2025}, pages 4257--4275, Albuquerque, New Mexico. Association for Computational Linguistics.

\bibitem[{Yadav et~al.(2025)Yadav, Suravee, Kemnitz, Strauss, and Yordanova}]{yadav-etal-2025-prompt}
Dipendra Yadav, Sumaiya Suravee, Stefan Kemnitz, Tobias Strauss, and Kristina Yordanova. 2025.
\newblock \href {https://aclanthology.org/2025.ranlp-1.158/} {Prompt engineering for {N}epali {NER}: Leveraging {H}indi-capable {LLM}s for low-resource languages}.
\newblock In \emph{Proceedings of the 15th International Conference on Recent Advances in Natural Language Processing - Natural Language Processing in the Generative AI Era}, pages 1364--1373, Varna, Bulgaria. INCOMA Ltd., Shoumen, Bulgaria.

\bibitem[{Zhang et~al.(2024)Zhang, Zhao, Gao, and Hu}]{LinkNER}
Zhen Zhang, Yuhua Zhao, Hang Gao, and Mengting Hu. 2024.
\newblock \href {https://doi.org/10.1145/3589334.3645414} {Link{NER}: Linking local named entity recognition models to large language models using uncertainty}.
\newblock In \emph{Proceedings of the ACM Web Conference 2024}, WWW '24, page 4047–4058, New York, NY, USA. Association for Computing Machinery.

\end{thebibliography}

\appendix

\section{Prompts}
\label{app:prompts}
\setlength{\parindent}{0pt}

\textbf{Without entity type definitions:} 

Your task is to identify named entities in Nepali tweets.
The entity types are Person (PER), Location (LOC), Organization (ORG), Event (EVENT), and Date (DATE).

EXAMPLES: \\
Input: \{example tweet\} \\
Output: \{list of mentions\} \\
...

Output your answer as JSON in the format
[["entity1", "type1"], ["entity2", "type2"], ...]
or [] if there are no entities. Return ONLY valid JSON and nothing else.

Input: \{test tweet\} \\
Output: \\

\textbf{With entity type definitions:}

Your task is to identify named entities in Nepali tweets.
The entity types are: \smallskip

1) Person (PER): Proper names of people including first names, family names, generational markers, nicknames, and fictional/mythological characters. Do not mark titles, relation names, pronouns, or name prefixes. \smallskip

2) Location (LOC): Man-made structures and politically defined places including buildings, cities, towns, city districts, continents, countries, states, geographical areas, parks, planets, celestial objects, and bodies of water. Do not mark generic references to locations or nationalities. \smallskip

3) Organization (ORG): Organizations including companies, commissions, communities / groups of people, education \& scientific institutes, judicial systems, law enforcement organizations, news agencies \& stations, political parties, public administrations, sport leagues \& clubs, and banks. \smallskip

4) Event (EVENT): Named events and phenomena including natural disasters, revolutions, battles, wars, demonstrations, concerts, and sports events. \smallskip

5) Date (DATE): Date or period of 24 hours or more including days, weeks, months, named periods, seasons, years, and ages. \smallskip

EXAMPLES: \\
Input: \{example tweet\} \\
Output: \{list of mentions\} \\
...

Output your answer as JSON in the format
[["entity1", "type1"], ["entity2", "type2"], ...]
or [] if there are no entities. Return ONLY valid JSON and nothing else.

Input: \{test tweet\} \\
Output: \\

\textbf{Self-verification:}

You are an expert at identifying named entities in Nepali and English.

The \{TYPE\} entity type is defined as: \{definition\}.

Consider this tweet: \{tweet\}

Is \{span\} a named entity of type \{TYPE\}? Answer "Yes" or "No".

\end{document}